\documentclass[sigconf]{acmart}
\AtBeginDocument{%
  \providecommand\BibTeX{{%
    \normalfont B\kern-0.5em{\scshape i\kern-0.25em b}\kern-0.8em\TeX}}}

\setcopyright{acmcopyright}
\copyrightyear{2022}
\acmYear{2022}
\acmDOI{XXXXXXX.XXXXXXX}

\acmConference[CONCATENATE - HRI]{}{March 13, 2023}{Stockholm, SE}
\usepackage{caption}
\usepackage{subcaption}
\begin{document}

\title{Measuring the perception of the personalized activities with CloudIA robot}
\author{Alessandra Sorrentino}
\email{alessandra.sorrentino@unifi.it}
\orcid{0000-0003-3187-810X}
\affiliation{%
  \institution{Department of Industrial Engineering}
  \streetaddress{Via Santa Marta 3}
  \city{University of Florence, Florence}
  \country{Italy}
}

\author{Laura Fiorini}
\affiliation{%
  \institution{Department of Industrial Engineering}
  \streetaddress{Via Santa Marta 3}
  \city{University of Florence, Florence}
  \country{Italy}}
\affiliation{%
  \institution{BioRobotics Institute, Scuola Superiore Sant'Anna}
  \streetaddress{Viale Rinaldo Piaggio 34}
  \city{Pontedera}
  \country{Italy}}

\author{Carlo La Viola}
\affiliation{%
  \institution{Department of Industrial Engineering}
  \streetaddress{Via Santa Marta 3}
  \city{University of Florence, Florence}
  \country{Italy}}

\author{Filippo Cavallo}
\affiliation{%
  \institution{Department of Industrial Engineering}
  \streetaddress{Via Santa Marta 3}
  \city{University of Florence, Florence}
  \country{Italy}}
\affiliation{%
  \institution{BioRobotics Institute, Scuola Superiore Sant'Anna}
  \streetaddress{Viale Rinaldo Piaggio 34}
  \city{Pontedera}
  \country{Italy}}

\renewcommand{\shortauthors}{Sorrentino, et al.}

\begin{abstract}
Socially Assistive Robots represent a valid solution for improving the quality of life and the mood of older adults. In this context, this work presents the CloudIA robot, a non-human-like robot intended to promote sociality and well-being among older adults. The design of the robot and of the provided services were carried out by a multidisciplinary team of designers and technology developers in tandem with professional caregivers. The capabilities of the robot were implemented according to the received guidelines and tested in two nursing facilities by 15 older people. Qualitative and quantitative metrics were used to investigate the engagement of the participants during the interaction with the robot, and to investigate any differences in the interaction during the proposed activities. The results highlighted the general tendency of humanizing the robotic platform and demonstrated the feasibility of introducing the CloudIA robot in support of the professional caregivers’ work. From this pilot test, further ideas on improving the personalization of the robotic platform emerged. 
\end{abstract}






\maketitle

\section{Introduction}
During the latest years, social assistive robots (SARs) have gained a growing interest due to promising results as therapy tools. Several studies investigate the effectiveness of robot interventions for supporting children’s mental well-being \cite{kabacinska2021socially} as well as for helping children with autism spectrum disorders during social therapies with professional caregivers \cite{martinez2020socially}. Similarly, SARs have been proposed as a valid solution for assisting older adults, both to prolong independent living \cite{bevilacqua2014telepresence} and to support the workload of professional caregivers within nursing home activities \cite{abdi2018scoping,pineau2003towards}. 

In the latter context, there have been several pieces of evidence that SARs have been perceived as trustworthy companions by the primary users and useful tools for fostering social interactions \cite{sharkey2012granny}. Namely, previous research proved that SARs could have a positive impact in reducing depression, loneliness, and isolation while promoting social and emotional interactions with older users \cite{stiehl2005design}. In this direction, pet-like robots, like PARO and NeCoRo, have been widely adopted for stimulating an emotional response as in human-animal interaction. The outcomes of these studies report improvements in quality of life, mood, and relationships in older adults who have interacted with robotic pets \cite{takayanagi2014comparison,valenti2015social}. Due to the recent advancements in interaction modalities, the latest studies focus on the adoption of human-like robots (e.g., NAO and Pepper robots). Due to their aspect, human-like robots facilitate social interaction and communication, as they possess all the necessary features to convey social signals \cite{desideri2019emotional}. Thanks to their social capability, they are mostly used as support tools for cognitive assessment and stimulation tasks \cite{fan2016robotic,manca2021impact,pino2020humanoid,rossi2018psychometric}, establishing dyadic interactions with older people. The advantages of adopting SARs are two-fold. On one side, they allow multiple screenings in a reduced time frame, guaranteeing the required standardization of the cognitive test \cite{di2019assessment}. Furthermore, this strategy could alleviate the workload of professional caregivers, by providing ongoing quantitative assessments \cite{sorrentino2022personalizing}. With respect to the technologies currently used in this context (e.g. tablet and PC), the presence of the robot results in a more engaging and empathetic interaction \cite{mann2015people}. 

This work proposes the introduction of a non-human-like robot as a support tool for professional caregivers in nursing facilities. The design and the tasks of the robot were identified in collaboration with three social cooperatives in the Tuscan territory (Italy), adopting the methodological approach of Human-Centred Design (HCD) and Ergonomics in Design for the home care sector, as described in \cite{sorrentino2022design}. Two social cooperatives were nursing homes hosting older people, the latter cooperative was an assisted living facility for young people with severe cognitive disabilities. This methodological approach gave us the opportunity to personalize the appearance as well as the services provided by the robot, to better fit the context and the target users. As result, the appearance of the robot was designed in a non-human-like fashion. With respect to the non-human-like robots proposed in \cite{luperto2019evaluating,sorrentino2021feasibility}, the robot has been endowed with more social skills required to conduct an interaction (i.e., facial and verbal expressions). Thus, the first aim of this work is to investigate whether, despite its appearance, the robot is perceived as a social actor in the interaction. 
The personalization of the services provided by the robot focused on the identification of the activities that the residents could perform with the robot, reducing the intervention and the working load of the professional caregivers. In the end, two application scenarios were identified: (i) administration of standard psycho-cognitive assessment tests, and (ii) encouraging socialization among users through music therapy. Since different scenarios and target users were involved, each service has been customized accordingly (e.g. type of music, selected test). Interestingly, both activities were already performed by the residents with professional caregivers. Thus, the second objective of this work is to explore whether the presence of the robot in these activities influences the interaction modalities and relationships among the participants, with respect to the traditional modality (i.e., only a professional caregiver is present). Since the two types of activities promote different interaction paradigms (i.e., human-robot interaction and humans-robot interaction), the third objective of this work is to examine if the level of social interaction with the robot established during the activities changes according to the type of activity. Qualitative and quantitative evaluation tools were adopted to provide an answer to these three research objectives. 

In our previous work \cite{sorrentino2022designEMBC}, we have already investigated these three aspects, based on the feedback received in the assisted residential facility context. In this work, we describe the results obtained by introducing the CloudIA robot in the second context, i.e. the two nursing facilities.  

\section{Material}
\subsection{CloudIA robot}
\begin{figure}[h]
  \centering
  \includegraphics[width=0.9\linewidth]{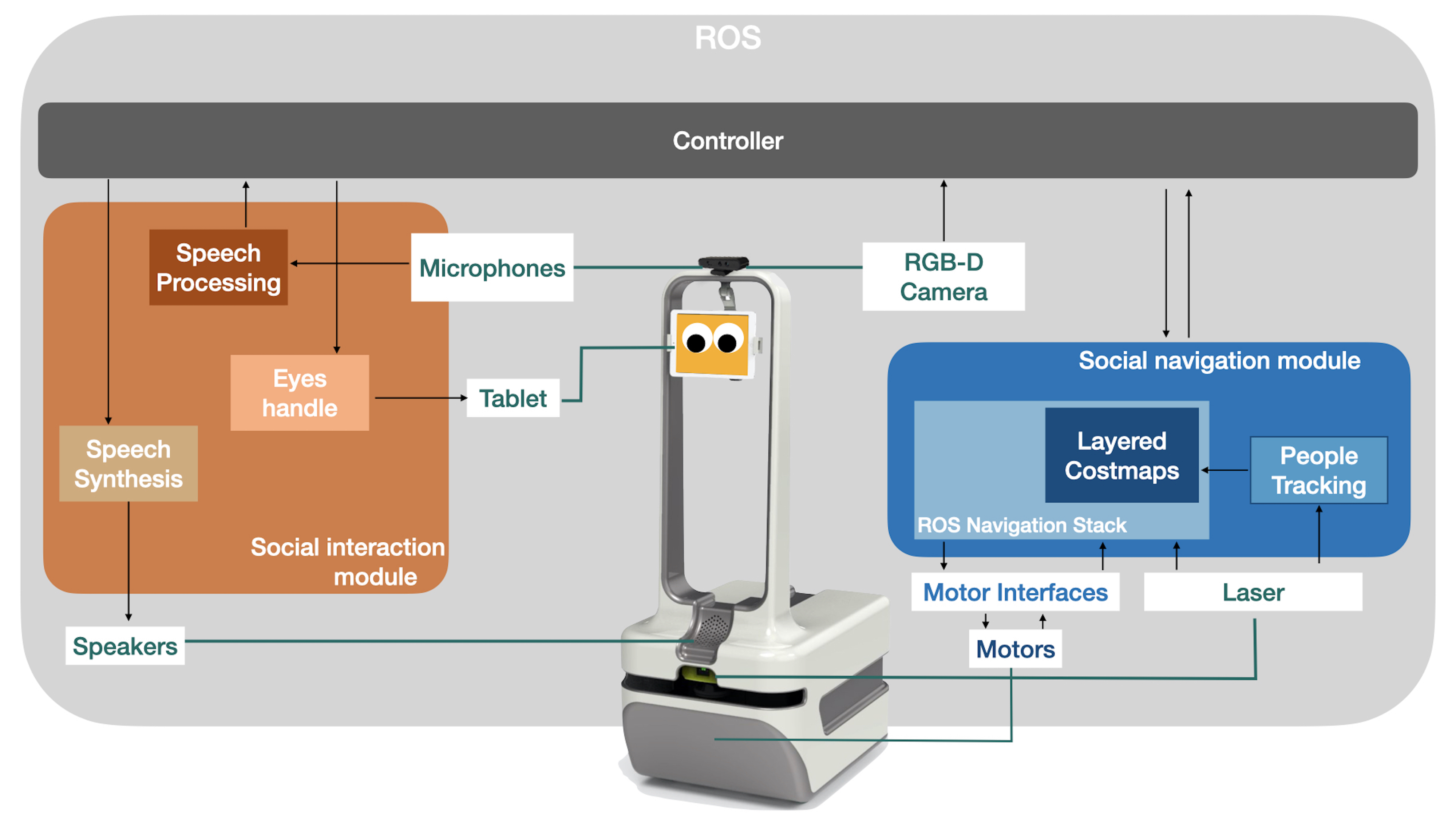}
  \caption{Hardware and software architecture of CloudIA robot.}
  \Description{Hardware and software architecture of CloudIA robot.}
  \label{fig:1}
\end{figure}

This work adopted a non-human-like robot named CloudIA robot. The appearance and the tasks of the robot have been designed based on Human-Centred Design and Ergonomics in Design methodologies, as described in \cite{pistolesi2019metodologie} and \cite{sorrentino2022design}, respectively. From a technical point of view, CloudIA is a ROS-based robot capable of autonomously navigating the environment\cite{sorrentino2021modeling}, and supports social skills. As shown in Figure \ref{fig:1}, the robot is equipped with an ASTRA PRO 3D camera (Orbbec, USA) on the top, two stereo sound speakers (Trust, USA), and a SICK TIM781 laser (SICK, Germany) on the bottom part of the robot’s base. A tablet (Samsung, South Korea) is located on the top part of the robotic platform, on which a graphical web interface displays the facial expression of the robot. In detail, we implemented two facial expressions: attentive (i.e., the eyes look straight, while blinking), and distracting behavior (i.e., the eyes randomly rotate in four directions). The color’s choice of facial expressions reflected the emotion of joy \cite{loffler2018multimodal}. The robot is endowed with speech capability. Namely, the svox-pico Text-to-Speech (TTS) engine, which is used as a speech synthesizer. 

\subsection{Activity description}
In conjunction with a team of experienced professional caregivers that work at the pilot site, the services delivered by the robot were selected. The idea was to implement some services that were currently part of the occupational therapy session so as to evaluate the acceptance of introducing a social robot. Then, the robot was programmed to execute some stimulation activities and routines supporting the staff psychologist. In the end, two main activities were implemented: 
\begin{itemize}
\item {cognitive assessment}: the role of the robot was to interact with the participant, asking to complete the tasks included in the Mini-Mental State Examination (MMSE \cite{folstein1983mini}), which aim to stimulate selected cognitive domains. The implementation of this activity followed the same strategy described in \cite{sorrentino2021feasibility}. This activity involved one individual at a time (i.e., individual activity).
\item{promote socialization through music-therapy}: the task of the robot was to foster a discussion among a group of older users based on the shared music listening’s experience \cite{agres2021music}. A list of songs was provided in advance by the caregivers and played directly by the robot when requested by the group. Once the song started, the robot started ``dancing'', by randomly moving in its place. Once ended, the robot asked for feedback about the song, fostering a discussion among the participants. Since multiple people were involved, it was identified as a group activity. 
\end{itemize}

\section{Case Study}
\subsection{Participants}
The activities with the robot were evaluated by two groups of users. Both groups included older individuals, living in a nursing home located in Tuscany, namely: RSA Uscita di Sicurezza (Orbetello, GR, Italy) and RSA Bonelle (Pistoia, PT, Italy). In both facilities, the professional caregivers were responsible for the recruitment of the participants among the residents. The inclusion criteria were residents who did not have hard hearing and/or speaking impairments, and those who have already participated in the same activities with the professional caregivers. Additionally, we included in the experimentation at least two professional caregivers for each facility. All the participants signed the consent form before entering the test. The study was approved by the joint Ethical Committee of Scuola Superiore Sant’Anna and Scuola Normale Superiore (Delibera 14/2020).

\subsection{Experimental protocol}
The experimental procedure adopted in this study is composed of mainly 4 stages. In the preliminary stage (T0, Welcoming), the participants were informed about the procedure and the capability of the CloudIA robot. During this stage, the robot was neither moving nor speaking. After the welcoming, the participants were firstly involved in the music-therapy activity (T1, Music), and then, in the cognitive assessment activity (T2, MMSE). In both activities, to accustom the participants to interact with the robotic platform, the robot asked the subject(s) some general questions (e.g., ``how are you?'', ``what is your name?''), before starting the planned activity. When the users were ready to start, the robot continued the interaction by proposing some pre-compiled questions or task competitions, based on the proposed activity. During the interaction, the robot could say some motivational sentences when triggered by the professional caregiver. At the end of the interaction, the robot always thanked the participant. During T1 and T2, a professional caregiver assisted the trial, supporting the participants during the interaction, if needed. At the end of the whole experimentation (T3, Leaving), the professional caregivers were contacted to conduct a brief interview about the overall experience. 

The experimentation adopted a Wizard-of-Oz strategy due to safety reasons, but also due to Covid-19 restrictions since the tests were performed in May 2021 and researchers were not allowed to enter the residential facility. Namely, the professional caregivers were instructed on how to use a web interface to guide the robot in performing the task (i.e., selecting the task/question of MMSE, and expressing encouragement). Just the facial expressions were set a priori: the attentive look was always active, except during the robot’s dance.

\section{Data collection and analysis}
The perception of the robot and the quality of interaction were evaluated in terms of the engagement of the participants during the activities with the robot. To this aim, the metrics used in this study rely on the direct observations of the participant’s behaviors during the interaction. We asked the professional caregivers involved in the experimentation to compile, during the T1 and T2, an observational grid adapted from \cite{cohen2009engagement}, for each participant. The observational grid focused on rating the level of interest shown by the participant (i.e. Interest domain), how active (i.e. Activeness domain) and attentive (i.e. Attention domain) the user was during the activity and the general attitude of the participant (i.e. Attitude domain) on a 7-Likert scale. Each domain is properly described in \cite{cohen2009engagement}. Descriptive statistics (i.e. median and interquartile range, IQR) were computed for the grid’s domains, clustering the ratings based on the activity they belonged to. Additionally, we asked the professional caregivers to annotate the factors they considered during the evaluation. 
To collect some additional comments on the experience, we asked the professional caregivers some general questions during T3, namely: (a) ``What was your impression of the robot experimentation?''; (b) ``Please, tell me about your perception of how the services were experienced by the participants. Which was the service that the participants liked the most?''; (c) ``Do you think the robot can be a support to your work? If yes, which is the task the robot should perform?''; (d) ``Did you notice any technical criticality related to the robot and/or the services?''. Qualitative results were obtained by analyzing the answers.

\section{Results}
\begin{figure}[h]
  \centering
 \begin{subfigure}[b]{0.3\textwidth}
 \centering
         \includegraphics[width=\textwidth]{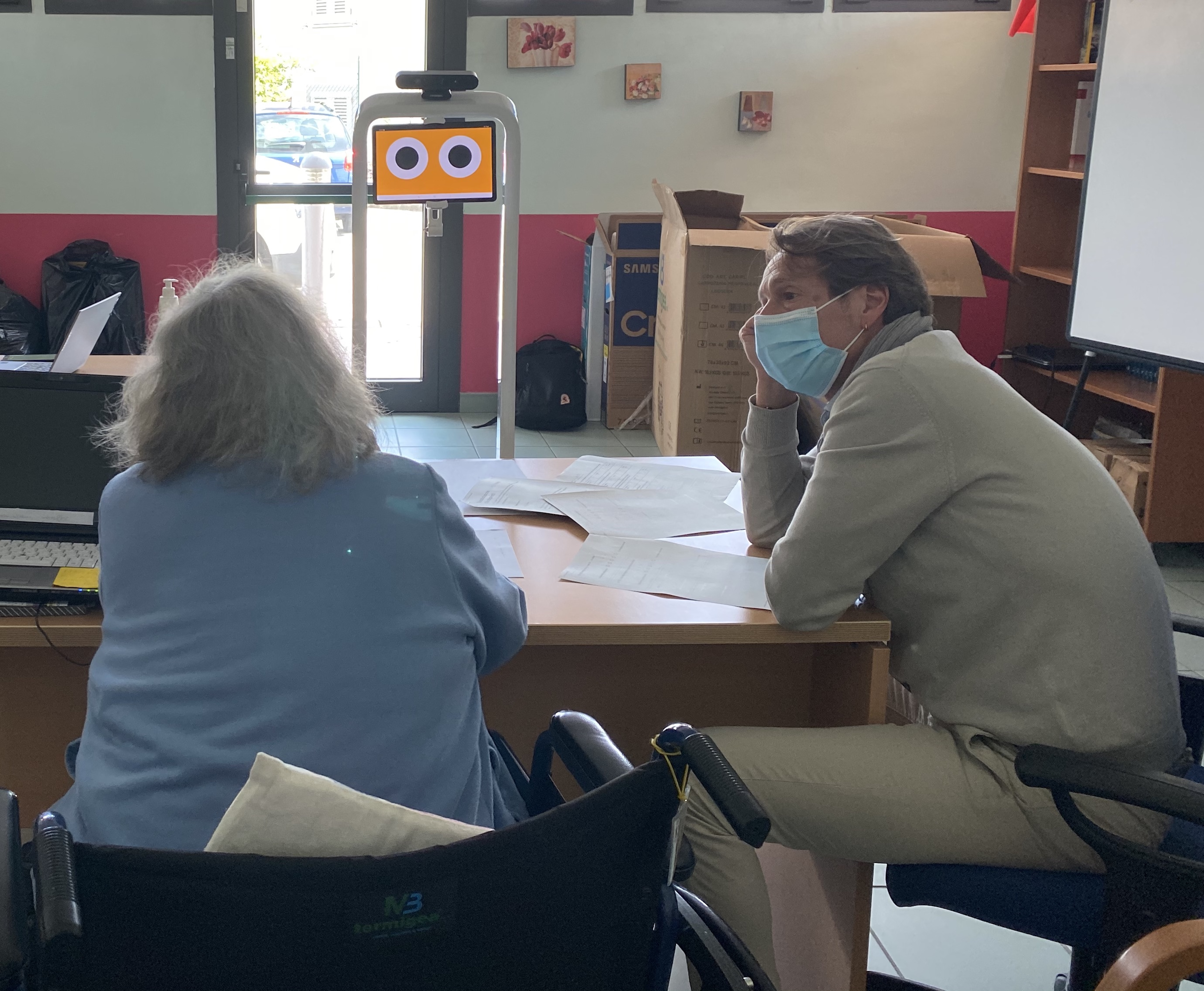}
         \caption{}
         \label{}
     \end{subfigure}
     \hfill
     \begin{subfigure}[b]{0.4\textwidth}
         \centering
         \includegraphics[width=\textwidth]{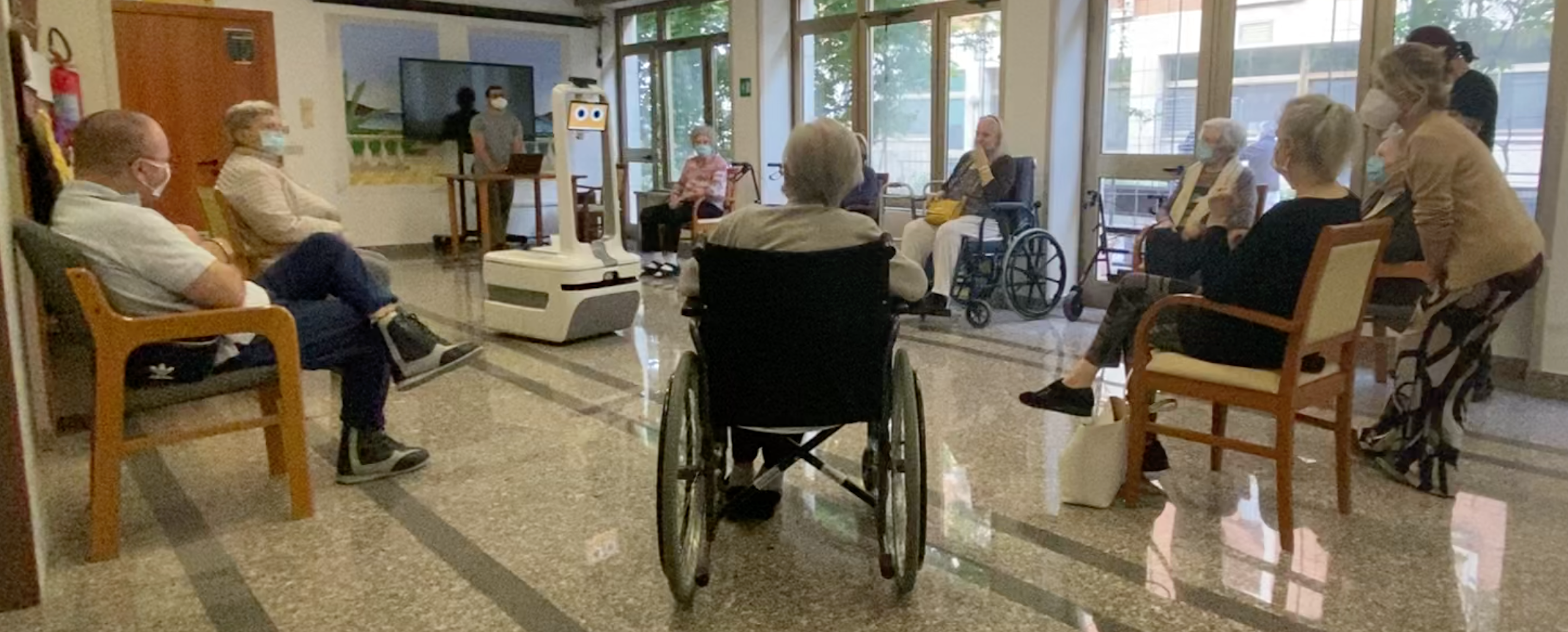}
         \caption{}
         \label{}
     \end{subfigure}
        \caption{Participants interacting with CloudIA robot, during (a) individual and (b) music-therapy activity.}
        \label{figure:es}
\end{figure}

A total of 15 older users (14 female and 1 male, avg age=80.9 years old; std age=$\pm$13.06) and 4 professional caregivers were involved in this case study. Two group sessions were organized, which involved 10 participants in one residential facility and 5 end-users in the other one. In the former residential facility, the group session lasted about 90 minutes, while, in the latter one, it lasted 30 minutes. Overall, 8 older users participated to the individual sessions, where they interacted with the CloudIA robot for 12.07 minutes on average (std=$\pm$3.05). While Figure \ref{figure:es}(a) shows a participant undergoing the cognitive assessment with the robot, Figure \ref{figure:es}(b) shows a group session.

\subsection{Quantitative results}
\begin{figure}[h]
  \centering
  \includegraphics[width=0.9\linewidth]{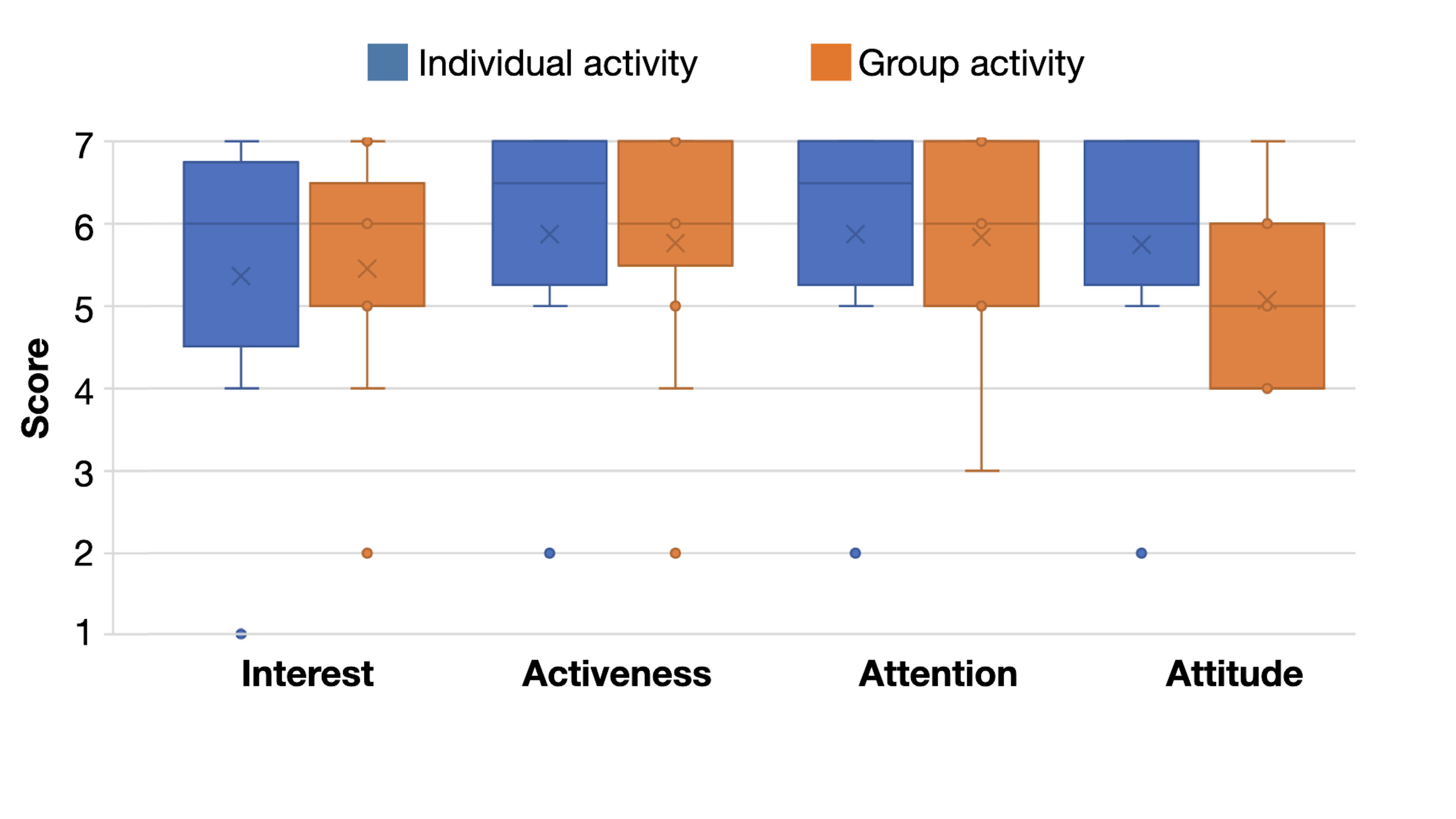}
  \caption{Box and Whisker plots of the engagement domains rated by the professional caregivers based on the participants' behaviors during the interaction.}
  \Description{Box and whisker plots of evaluation scores of individual and group activity, separately.}
  \label{figure:box}
\end{figure}

The engagement of the participants was evaluated in terms of interest, activeness, attention, and attitude during the interaction. We analyzed these aspects about the type of activity they were involved in (e.g., individual or group activity). Regarding the first domain, the participants appeared to be very interested and showed great participation in both individual (median=6; IQR=$\pm$0.75) and group (median=6; IQR=$\pm$1) activities. As shown in Figure \ref{figure:box}, the professional caregivers noticed that the participants were slightly more active and attentive during the individual activity (median=6.5; IQR=$\pm$1.25). Instead, in the group activity, the participants were active (median=6; IQR=$\pm$1), but some of them were not always attentive (median=6; IQR=$\pm$2). Similarly, the professional caregivers noted a general positive attitude of the participants involved in the interaction. Among the activities, the participants’ attitude was on average slightly more positive during the individual activity (median=6; IQR=$\pm$1.25) than during the group activity (median=5, IQR=$\pm$2). Among the factors that the caregivers considered for evaluating the behaviors of the participants during the interaction, they mostly mentioned psycho-emotional states of the participants in the individual activity, and the number of questions/answers expressed during the group activity.

\subsection{Qualitative results}
Analyzing the feedback of the professional caregivers at the end of the experimental sessions, it emerged they were positively and pleasantly surprised by the overall experience and the older people' reactions (answer Question (a)). Both caregivers and older people were very impressed by the minimal anthropomorphism of the robot, which nonetheless stimulated residents to interact with it (``\textit{all the guests expected a human-like robot. When they first saw the robot (it was turned off), they were all disappointed. Then, when the robot opened its eyes, it took on humanity and they liked it more}'', statement of one caregiver). They were very surprised by the ease by which the residents talked to the robot, treating it as if it were a human being: ``\textit{the robot was well received by the residents, there was no resistance. It seemed as if the residents wanted to establish an intimate relationship. The residents related to the robot as well as they would relate to a person}'' (statement of one caregiver);``\textit{although they linked the robot to a machine, they were all happy to respond well and with satisfaction}'' (statement of another caregiver). Regarding the activity that the participants liked the most (Question (b)), the professional caregivers agreed on selecting social promotion as the activity in which the participants felt more involved (``\textit{They were all actively involved in the experience}'', statement of one facility manager). The caregivers also remarked the fact that the robot emanated joy and happiness among the guests (``\textit{the experience was characterized by hilarity and fun}'', statement of caregiver). In addition, the professional caregivers pointed out that, during this activity, some participants also showed different behaviors than when they do the same activity with them (``\textit{When they perform this activity with me or my colleagues, they barely talk to each other. Instead today they helped each other, commenting the other people song’s choices as well as the robot’s abilities}'', statement of caregiver). Regarding the cognitive assessment activity, caregivers stated that, given the nature of the interaction (one-to-one interaction with considerable cognitive effort), and the occurrence of the robot’s voice-related problems, many of the participants felt guilty when they did not hear the robot properly or when they did not know the correct answer. From the professional caregivers' perspective, the guilt sensation presupposes a ``\textit{desire to look good in front of the robot}'' (statement of the caregiver). Of the cognitive stimulation service, users appreciated the motivational and comforting phrases. In fact, ``\textit{participants felt proud and happy to correctly answer the robot}'' (statement of caregiver). 

Regarding the usability of the robot in this scenario (Question (c)), the professional caregivers agreed on using it as a support tool for their work, especially for promoting socialization. The caregivers recognized the potential of the robot both to positively distract residents from their daily routine and to become a ``listening point'', where older people can talk about their day or their state of mind. It is because ``\textit{the robot has been a diversion from daily life. For example, today it helped a person who spends all day checking on the phone and waiting the call of her son, to not think about it during the interaction with the robot. For the 15-20 minutes that she was engaged in dialogue with the robot, she didn't think about it}'' (statement of the caregiver). Many of the professional caregivers highlighted the limited autonomy of the robot (Question (d)), which requires a constant presence of a person. In addition, the professional caregivers highlighted the problems related to the robot's voice, which was not always understandable during the interaction, suggesting of moving the position of the speakers to a higher position and/or adding a mouth on the robot's screen to help older users to catch the moment when the robot speaks. Some professional caregivers also stated that the size of the images shown on the tablet was too small in the MMSE administration, suggesting that they should be enlarged or that the robot should be equipped with a larger screen. Aside from suggestions to improve the experience with the robot, the caregivers envisioned long-term use of the robot, suggesting some improvements for its actual use in the facilities, like the personalization of the services based on the patient (i.e.  ``\textit{The robot's behavior cannot always be the same. Its capabilities should change to stimulate the guests' attention and interest. [...] It would be interesting if the robot would update its list of songs according to the resident’s taste}'', statement of the facility manager).
The complete list of suggested improvements are reported in Table \ref{tab:freq2}. 

\begin{table}
  \caption{Guidelines for improving design and behaviors of CloudIA robot.}
  \label{tab:freq2}
  \begin{tabular}{cl}
    \toprule
    \textbf{Aspect}&\textbf{Guidelines}\\
    \midrule
    Design & Locate the speakers at a higher position.\\
     & Increase the dimension of the robot’s screen.\\
     \hline
    Interaction & Improve the quality of the robot’s voice.\\
    modality & Mimic the lips’ motion when the robot speaks.\\
     \hline
    Robot's & Motivate the user during the task.\\
    behaviors & Customize services based on user and context.\\
    & Guarantee autonomy in executing the tasks.\\
  \bottomrule
\end{tabular}
\end{table}

\section{Discussion}
Overall, the outcomes of this case study validated the introduction of the non-human-like robot CloudIA as support tool in nursing facilities. From the behavioral analysis of the 15 older participants and from the feedback expressed by the professional caregivers, the presence of the robot in this context enriched the interaction experience of the older participants. 

From an aesthetic perspective, the experience of the participants was characterized by some disillusionment. Most of them expected a human-like robot (i.e., with legs and arms) and, instead, they met a robot that looked more like a machine. All the professional caregivers involved in the study mentioned the concept of ``personification'', as the capability of treating the robot as if it was a person. The personification was the aspect that surprised the caregivers the most. It suggests that the end-users associated a much higher ability to reason than human appearance.  This leads us to conclude that although the robot did not have human-like appearance, its ability to properly ask/answer questions, and to move among the participants ``\textit{transformed the robot from an object to a subject}'' (caregiver’s statement). Additionally, the professional caregivers noticed that the participants interacted with the robot as if it was a person, e.g. by looking at it during the interaction and naturally answering to its questions. Similar results were also described in \cite{manca2021impact}, where the authors noticed that older adults adopted human-like social behaviors while interacting with Pepper robot (i.e. touching the hands and arms of the robot). In our study, older people interacted with a non-human-like robot, and it suggests that no matter the appearance, its presence incentive social behaviors. Differently from the work described in \cite{sorrentino2021feasibility}, in which the participants adopted social conventions with the non-human-like robot ASTRO during a cognitive assessment, our study extends the same results to a different type of activity, suggesting that the presence of other people did not influence the interaction established between one user and the robot. On the contrary, the presence of the robot (and the performed activity) positively influenced the interaction modalities among the participants, fostering a discussion. A similar result also emerged when CloudIA delivered the same services in the assisited residencial facility, as described in \cite{sorrentino2022designEMBC}. Namely, the participants associated human characteristics to the robot, "filling the absence  of the body with imagination" \cite{sorrentino2022designEMBC}.

From the comparison of the traditional modality and the presence of the robot during the activities (i.e. second objective), the professional caregivers noticed a great improvement during the group activity. In fact, they were positively surprised that group activity not only was characterized by jokes and hilarity, but it actually stimulated interaction among the participants, fulling the goal of the activity. Similarly, during the cognitive assessment, the presence of the robot also had a positive effect. In fact, the direct interaction with the robot distracted one participant from the daily routine of checking the phone, and it was commented on as a big achievement by the professional caregivers.  

Regarding the latter objective (i.e. examine if the level of social interaction with the robot established during the activities changes according to the type of activity), the professional caregiver noticed that the level of social interaction with the robot established during the activities changed according to the type of activity. During the individual activity, the participants appreciated the fact that they were the focus of attention of the robot. Additionally, the robot's ability to utter motivational and comforting phrases sometimes soothed the mood of a user in distress undergoing to a cognitive test. Being the focus of the robot's attention and receiving these encouragements positively affected the final mood of the participants. It explains why the overall attitude of the participants was slightly more positive during cognitive stimulation than during the social promotion service. Since in the group activity, the robot shifted the focus of attention among the participants, it may also explain why the engagement of the participants in the group activity was characterized by higher variability: lower engagement when the robot was speaking with other participants, and vice-versa. 

The introduction of CloudIA in two different environments, hosting people of different ages and cognitive disabilities, returned similar feedback. In both cases, the residents associated human characteristics with the robot, ``filling the absence  of the body with imagination'' \cite{sorrentino2022designEMBC}, and the presence of the robot stimulated social behaviors during the interaction, positively affecting the mood of the participant during the activity. 

From the direct use of the platform, some critical issues as well as new ideas for improving the existing platform emerged (as reported in \ref{tab:freq2}). In future works, we would like to address these issues, improving the autonomy of the robot at the interaction level and adapting its behavior to guarantee long-term and successful interactions with older people.

\begin{acks}
This research received funding from the European Union’s Horizon 2020 research and innovation program under Grant Agreement No. 857188 (Pharaon Project).

The authors would like to thank Prof. Francesca Tosi, Mattia Pistolesi, and Claudia Becchimanzi for their valuable work in designing the robotic platform. Furthermore, the authors sincerely thank the directors, the professional caregivers, and the residents of RSA Uscita di Sicurezza (Orbetello, GR, Italy) and of RSA Bonelle (Pistoia, PT, Italy), where the experimentation took place, for their availability and enthusiasm in designing and testing the activities with CloudIA. 
\end{acks}


\end{document}